%% file: main.tex
\documentclass[10pt,twocolumn,letterpaper]{article}

\usepackage{cvpr}              

\usepackage{graphicx}
\usepackage{amsmath}
\usepackage{amssymb}
\usepackage{booktabs}
\usepackage{multirow}
\usepackage{cuted}
\usepackage{bbding}
\usepackage{makecell}
\usepackage[accsupp]{axessibility} 

\usepackage[pagebackref,breaklinks,colorlinks]{hyperref}

\usepackage[capitalize]{cleveref}
\crefname{section}{Sec.}{Secs.}      
\Crefname{section}{Section}{Sections}
\Crefname{table}{Table}{Tables}
\crefname{table}{Tab.}{Tabs.}
\def\cvprPaperID{6557} 
\def\confName{CVPR}
\def\confYear{2022}
\begin{document}
\title{Learning a Structured Latent Space for Unsupervised Point Cloud Completion}
\author{Yingjie Cai\textsuperscript{1}, 
\ Kwan-Yee Lin\textsuperscript{1}\thanks{H. Li and K. Lin are the co-corresponding authors.},
\ Chao Zhang\textsuperscript{2},
\ Qiang Wang\textsuperscript{2},
\ Xiaogang Wang\textsuperscript{1},
\ Hongsheng Li\textsuperscript{1}\footnotemark[1] \\
\textsuperscript{1}CUHK-SenseTime Joint Laboratory, The Chinese University of Hong Kong \\
\textsuperscript{2}Samsung Research Institute China - Beijing (SRC-B) \\
{\tt\small caiyingjie@link.cuhk.edu.hk}
}
\maketitle

\begin{abstract}
Unsupervised point cloud completion aims at estimating the corresponding complete point cloud of a partial point cloud in an unpaired manner. It is a crucial but challenging problem since there is no paired partial-complete supervision that can be exploited directly. In this work, we propose a novel framework, which learns a unified and structured latent space that encoding both partial and complete point clouds. Specifically, we map a series of related partial point clouds into multiple complete shape and occlusion code pairs and fuse the codes to obtain their representations in the unified latent space. To enforce the learning of such a structured latent space, the proposed method adopts a series of constraints including structured ranking regularization, latent code swapping constraint, and distribution supervision on the related partial point clouds.
By establishing such a unified and structured latent space, better partial-complete geometry consistency and shape completion accuracy can be achieved. Extensive experiments show that our proposed method consistently outperforms state-of-the-art unsupervised methods on both synthetic ShapeNet and real-world KITTI, ScanNet, and Matterport3D datasets.
\end{abstract}
\section{Introduction}
Point cloud completion aims at estimating the corresponding complete point cloud of a partial point cloud, which is an important task and can assist downstream applications such as shape classification~\cite{qi2017pointnet, han20193dviewgraph, liu2021fine, liu2020lrc, liu2019l2g}, robotics navigation~\cite{engel2014lsd, mur2015orb} and scene understanding~\cite{dai2018scancomplete, cai2021semantic, hou20193d, cai2020monocular}, as raw point clouds are often noisy, sparse and partial.
Although fully supervised point cloud completion methods~\cite{hu20203d, xie2020grnet, yin2018p2p, huang2020pf,yuan2018pcn, liu2020morphing,wen2021pmp, yu2021pointr, xiang2021snowflakenet} have achieved impressive performance, they heavily rely on large-scale paired partial-complete training data. It is, however, difficult to collect paired data from real-world scans. Additionally, such completion networks trained on paired real data or paired synthetic data cannot sufficiently generalize to actual scans, as their data distributions might not well match those of training samples.
\begin{figure}[t!]
\centering
\includegraphics[width=0.9\linewidth]{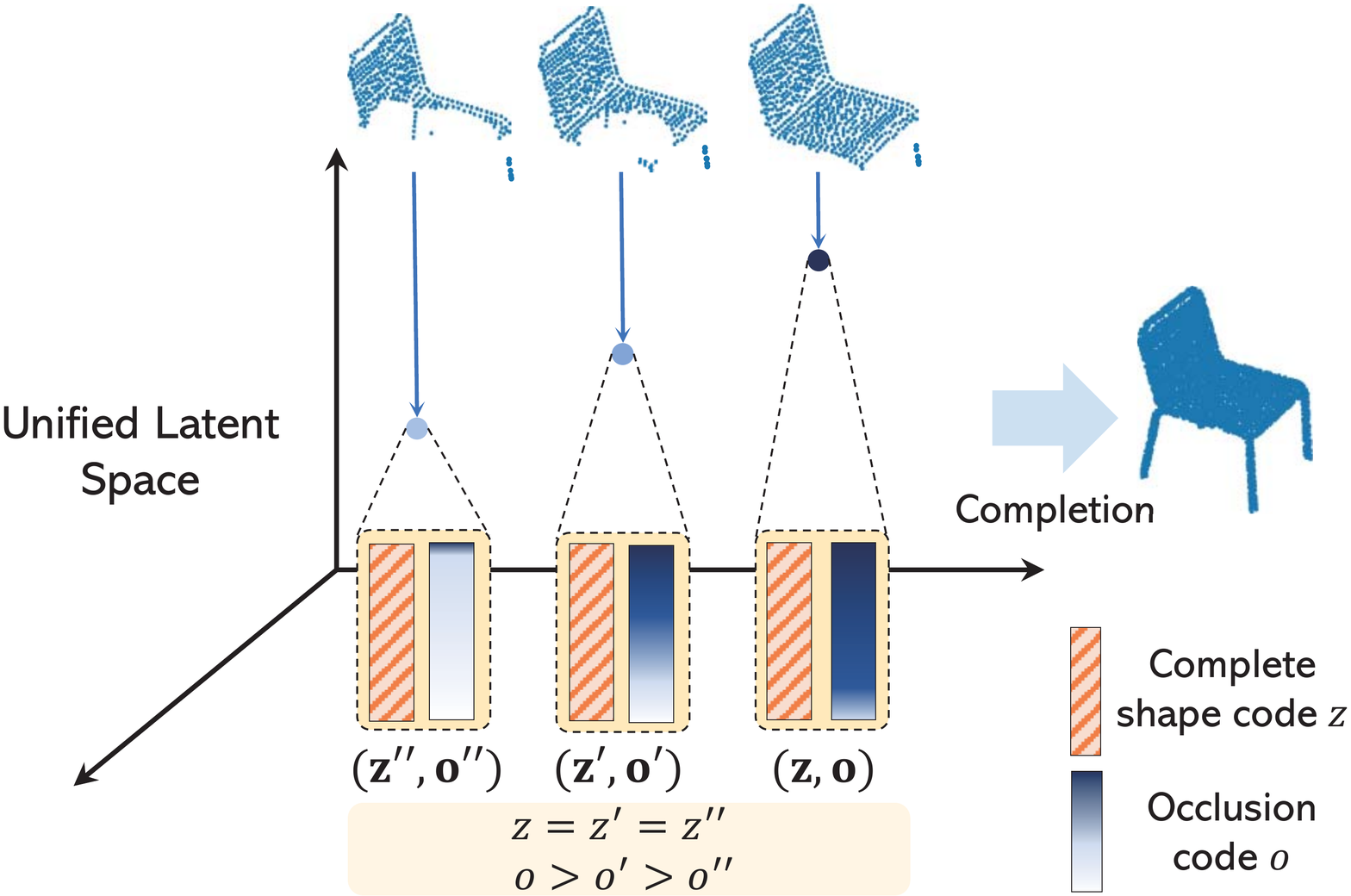}
\caption{Illustration of the unified and structured latent space where any point cloud can be represented as a complete shape code $\mathbf{z}$ and a corresponding occlusion code $\mathbf{o}$. We complete the partial point cloud via properly regularizing such codes in an unsupervised manner. Best viewed in color.}
\label{fig:fig1}
\vspace{-0.4cm}
\end{figure}

A promising alternate solution is to learn a completion network in an unpaired manner following the setup of~\cite{chen2019unpaired, wu2020multimodal, zhangunsupervised,wen2021cycle4completion}. However, it is a more challenging setup since there is no paired and accurate point-wise supervision that can be adopted directly. 
To tackle the problem, different methods are proposed to adopt different types of supervision from the unpaired data.
A representative work~\cite{zhangunsupervised} adopts GAN inversion for 3D shape completion. It trains a complete point cloud generator with adversarial losses. During inference, an optimal shape code can be recovered via hundreds of gradient descent iterations by minimizing a partial-complete consistency loss. 
The consistency\footnote{Consistency describes whether the predicted point cloud represents the same object as the partial input.} between the predicted point cloud and the input can be maintained. However, the inverse optimization is generally unstable and easy to stuck at local minima if using unsuitable initial code, unsuitable learning rate or too many iterations, etc. The inversion process is much more time-consuming than direct methods ($\sim$3500x).
Another representative unsupervised work~\cite{wen2021cycle4completion} exploits cycle supervisions to enhance consistency indirectly by learning bidirectional transformations between the latent spaces of complete and incomplete shapes (point clouds). 
However, the bidirectional transformations need to be separately modeled and are difficult to learn, especially for the complete-to-partial mapping. If one direction is not learned well, the other direction would be influenced correspondingly. 
In summary, without direct and accurate paired supervision, designing proper supervision and applying to unsupervised point cloud completion is of great importance to this task.
\begin{figure*}[ht!]
\centering
\includegraphics[width=\textwidth]{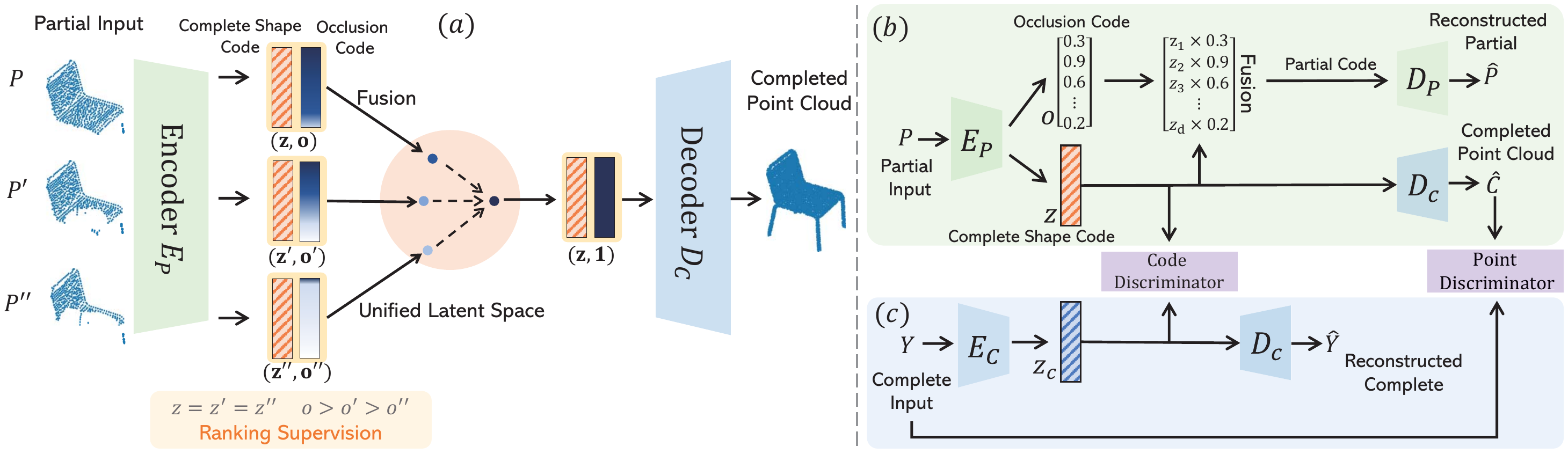}
\caption{\textbf{Overview.} (a) A series of related partial point clouds encoded to multiple complete shape and occlusion code pairs. Their element-wise multiplication are the representation in the unified latent space. (b) Reconstructing the partial input $\hat{P}$ and predicting the completed point cloud $\hat{C}$ simultaneously with a shape latent code discriminator and a complete point cloud discriminator. (c) The real complete shape codes and point clouds are provided by the complete point cloud auto-encoder. Best viewed in color. }
\label{fig:framework}
\vspace{-0.4cm}
\end{figure*}

To this end, we propose to create a unified and structured latent space for encoding both partial and complete shapes. 
To apply strong supervisions for unsupervised point cloud completion, we make an assumption that 
each partial shape is created by occluding a complete one.
If a complete shape is occluded to become partial in the 3D space, its code in the latent space should also be ``occluded" from a complete shape code accordingly. We model the ``occlusion" of a complete code in the latent space as weighting each of its dimension with a weight in $\left[0,1\right]$. However, instead of manually determining the occlusion weights, we make them learned from the training data. In this way, the complete and partial shapes are strongly bounded in a unified latent space.
In addition, to better regularize the relation between partial point clouds from the same complete shape, the occlusion code of a more occluded shape is required to have smaller weights than that of a less occluded shape.

Specifically, to learn the unified latent space, we represent any partial or complete point cloud by two codes: a complete shape code and an occlusion code. 
The complete shape code can be fed into a completion decoder to reconstruct the corresponding complete point cloud. 
The ``occluded'' shape code via multiplying the above two codes can be fed into a partial decoder to reconstruct the partial shape. 
Furthermore, we create a series of related partial point clouds by gradually removing more points from a partial shape and apply ranking constraints to their occlusion codes by N-pair loss~\cite{sohn2016improved} according to their relative occlusion degrees. Their complete shape codes are required to be equal since they represent the same object. By adopting such properly designed strong supervisions, more accurate complete point clouds with better geometric consistency and shape details can be reconstructed.

We experiment on popular point cloud completion benchmarks, including a synthetic dataset (ShapeNet~\cite{Kopaczka2019}) and real datasets (KITTI~\cite{Geiger2012CVPR}, ScanNet~\cite{dai2017scannet} and Matterport3D~\cite{Matterport3D}). The proposed method outperforms state-of-the-art unsupervised methods~\cite{chen2019unpaired,zhangunsupervised,wen2021cycle4completion,wu2020multimodal}. Our main contributions are summarized as follows:
\begin{itemize}
    \item We propose to learn a unified and structured latent space for unsupervised point cloud completion, which encodes both partial and complete point clouds to improve partial-complete geometry consistency and lead to better shape completion accuracy. 
    
    \item We propose to constrain the complete and occlusion codes of a series of related partial point clouds to enhance the learning of the structured latent space.
    
    \item Experimental results demonstrate the superiority of the proposed method over state-of-the-art unsupervised point cloud completion methods on both synthetic and real datasets.
\end{itemize}
\section{Related Work}
\noindent \textbf{Point Cloud Completion.~}
Point cloud completion has played an important role for many downstream applications such as robotics~\cite{engel2014lsd, mur2015orb} and perception~\cite{dai2018scancomplete, hou2019sis, cai2020monocular, cai2021semantic}, which has seen significant development since the pioneering work PCN~\cite{yuan2018pcn} proposed. Most existing approaches like~\cite{dai2017shape,wang2020cascaded, huang2020pf,chen2019unpaired, mo2019partnet, wen2020point,liu2020morphing,tchapmi2019topnet, zhang2020detail,xie2020grnet,huang2020pf, yu2021pointr} are trained in a fully-supervised manner. Although supervised point cloud completion methods have achieved impressive results, they are difficult to generalize to real-world scans, since the paired data is difficult to collect for actual scans and their data distributions might not match well.
Pcl2pcl~\cite{chen2019unpaired} first proposes to complete the partial shapes in an unsupervised manner without the need for paired data, which trains two separate auto-encoders, for reconstructing complete shapes and partial ones respectively and learns a latent code transformation from the latent space of partial shapes to that of the complete ones. 
Its subsequent work~\cite{wu2020multimodal} outputs multiple plausible complete shapes from a partial input.
Base on Pcl2pcl, Cycle4completion~\cite{wen2021cycle4completion} exploits an extra complete-to-partial latent spaces transformation in addition of partial-to-complete direction to capture the bidirectional geometric correspondence between incomplete and complete shapes. Another unsupervised work ShapeInversion~\cite{zhangunsupervised} proposes to apply GAN inversion which utilizes the shape prior learned from a pre-trained generator to complete the partial shape in an unsupervised manner. However, the inverse optimization process is time-consuming compared with forward-based methods and the results are easy to stuck at local minima, which greatly limits the practical application of inversion-based methods. Different from existing methods, we propose to learn a unified latent space supervised by tailored structured latent constraints to reconstruct better complete shapes.

\noindent \textbf{Structured Ranking Losses.~}
Deep metric learning plays an important role in various applications of computer vision, such as image retrieval~\cite{movshovitz2017no,sohn2016improved}, clustering~\cite{hershey2016deep}, and transfer learning~\cite{oh2016deep}. 
The loss function is one of the essential components in successful deep metric learning frameworks and a large variety of loss functions have been proposed. 
Contrastive loss~\cite{chopra2005learning,hadsell2006dimensionality} captures the relationship between pairwise data points, i.e., similarity or dissimilarity. Triplet-based loss is widely studied~\cite{cui2016fine,schroff2015facenet,wang2014learning} and composed of an anchor point, a positive data point, and a negative point and aims to pull the anchor point closer to the positive point than to the negative point by a fixed margin $\delta$.
Inspired by this, recent ranking-motivated approaches~\cite{law2017deep,movshovitz2017no,schroff2015facenet,sohn2016improved,oh2017deep,oh2016deep} have proposed taking into consideration richer structured information across multiple data points and achieve impressive performance. Different from triplet loss who considers one negative point, N-pair loss~\cite{sohn2016improved} aims to identify one positive example from multi negative examples. 
\section{Method}
The goal of our work is to reconstruct the complete point cloud from an input partial point cloud with only unpaired data. Designing proper and strong supervisions is of great importance to tackle this challenging problem. We propose to learn a unified latent space for encoding both complete and partial point clouds (shapes). We first introduce the unified latent space in Section~\ref{section:framework}, which encodes complete and partial point clouds in a joint space. Then structured latent supervisions of a series of related partial point clouds are adopted to further regularize the learning of the space (Section~\ref{section:supervision}). The overall architecture is depicted in Figure~\ref{fig:framework}.

\subsection{A Unified Latent Space for Point Cloud Encoding}
\label{section:framework}
\vspace{-0.1cm}
We introduce the unified latent space to establish the relations between partial and complete point clouds in an unpaired manner. Partial point clouds can be considered as being created by occluding the complete shapes. A partial point cloud represents the same object as its corresponding complete point cloud and the difference between the complete and partial point is just their occlusion degrees, as shown in Figure~\ref{fig:framework} (a). Therefore, we embed the incomplete and the complete point clouds into a unified latent space equipping with different occlusion degrees.

Specifically, as illustrated in Figure~\ref{fig:framework} (b), we map any partial point cloud $P$ into a complete shape code $\mathbf{z}\in\mathbb{R}^d$ and a corresponding occlusion code $\mathbf{o}\in\mathbb{R}^d$ via a point cloud encoder $E_p$~\cite{Xie_2021_CVPR} consisting of EdgeConv~\cite{wang2019dynamic} layers. Each entry $z_i$, $i\in[1,\dots, d]$ of the occlusion code is bounded in $\left[0,1\right]$ by a sigmoid function and has the same length as the complete shape code. Occlusion of the complete shape in the latent space is modeled as softly ``gating'' each dimension of the complete shape code. A smaller occlusion value denotes more occlusion to a complete shape. The embedding of the partial shape in the unified latent space can then be obtained by element-wise multiplication of the two codes. The complete and partial codes are then fed into two separate decoders, $D_c$ and $D_p$, to generate completed point cloud $\hat{C}$ and reconstruct the input partial point cloud $\hat{P}$, respectively. The two separate decoders adopt the same architecture made up of a multiple layers perception (MLP) following~\cite{wen2021cycle4completion}. Both $\hat{C}$ and $\hat{P}$ are supervised by a point-wise Chamfer Distance (CD) loss with respect to the partial input. The point-wise reconstruction loss is expressed as:
\begin{equation}
\label{eq:rec}
\begin{aligned}
\mathcal{L}_{rec} = \mathcal{L}_{CD}(P, \hat{P}) + \mathcal{L}_{CD}(P, Deg(\hat{C})).
\end{aligned}
\end{equation}
For $(P, \hat{C})$, the bi-direction Chamfer Distance cannot be utilized directly, but only the Unidirectional Chamfer Distance (UCD) cannot provide enough supervision for the inference of the missing parts, so we degrade ($\ie$ $Deg$) the $\hat{C}$ into a partial point cloud following~\cite{zhangunsupervised}'s degradation module, where only top-\textit{k} nearest points with respect to partial point cloud are kept. 
In order to further encourage the predicted complete point clouds to represent reasonable shapes, a point cloud discriminator is adopted. We formulate the point cloud discriminator with WGAN-GP~\cite{gulrajani2017improved} loss as
\begin{equation}
\label{eq:D_code}
\mathcal{L}_{D}^{p}=\mathbb{E}_{\hat{C}} D(\hat{C})-\mathbb{E}_{Y} D\left(Y\right)+\lambda_{gp} \mathcal{T}_{D},
\end{equation}
where $\lambda_{gp}$ is a pre-defined weight factor and $\mathcal{T}_{D}$ is gradient penalty term, denoted as
\begin{equation}
\mathcal{T}_{D}=\mathbb{E}_{\hat{C}}\left[\left(\left\|\nabla_{{\hat{C}}} D(\hat{C})\right\|_{2}-1\right)^{2}\right].
\end{equation}
The code adversarial training loss for the encoder $E_p$ and decoder $D_c$ is
\begin{equation}
\label{eq:G_code}
\mathcal{L}_{G}^{p}=-\mathbb{E}_{\hat{C}} D(\hat{C}).
\end{equation}
Note that, during inference, only the encoder $E_p$ and decoder $D_c$ are needed.
\subsection{Structural Regularization of the Unified Space}
\label{section:supervision}
To further regularize the learning of the structured latent space, we create a series of related partial point clouds and propose several properly designed latent code supervisions, including structured ranking regularization, latent code swapping constraint, and latent code distribution supervision to enhance the learning of the structured latent space.
Specifically, given a partial input $P$, we can create a series of related partial point clouds (see $\{P, P', P''\}$ in Figure~\ref{fig:framework} (a)) by gradually removing more points. For $P'$ and $P''$, there are $K$ and $2K$ points removed from the initial partial shape $P$. Therefore, for a triplet of such related partial point clouds S=$\{ P, P', P''\}$, their occlusion degrees gradually increase. 

\noindent \textbf{Structured Ranking Regularization.~}
For their complete shape codes, since the point clouds represent the same object, their complete shape latent codes, $\mathbf{z}$, $\mathbf{z'}$, $\mathbf{z''}$ are required to be equal. And we adopt \textit{Smooth $L1$} loss to constrain them
\begin{equation}
\label{eq:equal}
L_{z}=L_{1}\left(\mathbf{z}, \mathbf{z'}\right) + L_{1}\left(\mathbf{z}, \mathbf{z''}\right).
\end{equation}
Furthermore, since their occlusion codes represent the increasing degree of occlusion, their corresponding occlusion codes' weights shall be smaller as their occlusion degrees increase. Such relations between their occlusion codes can be expressed as
\begin{equation}
\label{relative}
{o}_i^{\prime\prime} \leq {o}_i^{\prime} \leq {o}_i \leq 1  \quad  \quad \text{for} \quad i=1, \cdots, d,
\end{equation}
where $o_i^{\prime\prime}$, $o_i^{\prime}$, $o_i$ are the $i$th entry of the occlusion codes for $P''$, $P'$, $P$, respectively. To implement such a constraint, we adopt the N-pair loss~\cite{sohn2016improved} to constrain the proposed relative relations in Eq.~(\ref{relative}),
\begin{equation}
\label{eq:n-pair}
\begin{aligned}
L\left(\mathbf{a}, \mathbf{p}, \left\{\mathbf{n}_j\right\}_{j=1}^{N}\right)=\log (1+\sum_{j=1}^{N} \exp \left(\mathbf{a}^\top\mathbf{n}_j-\mathbf{a}^\top\mathbf{p}\right)).
\end{aligned}
\end{equation}
For the N-pair loss, there are one anchor sample $\mathbf{a}\in\mathbb{R}^d$, one positive sample $\mathbf{p}\in \mathbb{R}^d$ and $N$ negative samples $\mathbf{n}_j\in \mathbb{R}^d$, as shown in Eq.~(\ref{eq:n-pair}). 
By minimizing the loss function, the positive sample would be pulled closer to the anchor, and negative samples are pushed farther away from the anchor.
Here, we adopt N-pair loss by choosing different occlusion codes as anchors, which can be written as the following sets, where $\mathbf{1}\in\mathbb{R}^d$ is an all-one vector:
\begin{equation}
\label{set}
\begin{aligned}
&\left\{\mathbf{a}=\mathbf{1}, \quad \mathbf{p}=\mathbf{o}, \quad\left\{\mathbf{n}_{j}\right\}_{j=1}^{N}=\left\{\mathbf{o}^{\prime}, \mathbf{o}^{\prime \prime}\right\}\right\}, \\
&\left\{\mathbf{a}=\mathbf{o}, \quad \mathbf{p}=\mathbf{o}^{\prime}, \quad\left\{\mathbf{n}_{j}\right\}_{j=1}^{N}=\left\{\mathbf{o}^{\prime \prime}\right\}\right\}, \\
&\left\{\mathbf{a}=\mathbf{o}^{\prime \prime}, \quad \mathbf{p}=\mathbf{o}^{\prime}, \quad\left\{\mathbf{n}_{j}\right\}_{j=1}^{N}=\{\mathbf{o}\}\right\},
\end{aligned}
\end{equation}
Therefore, the proposed relative ranking relation Eq.~(\ref{relative}) between occlusion codes are constrained via applying the N-pair loss on each of the sets in Eq.~(\ref{set}).
Through adopting such strong ranking regularization, the unified latent space is trained to be more structured.
\begin{figure}[t!]
\centering
\includegraphics[width=\linewidth]{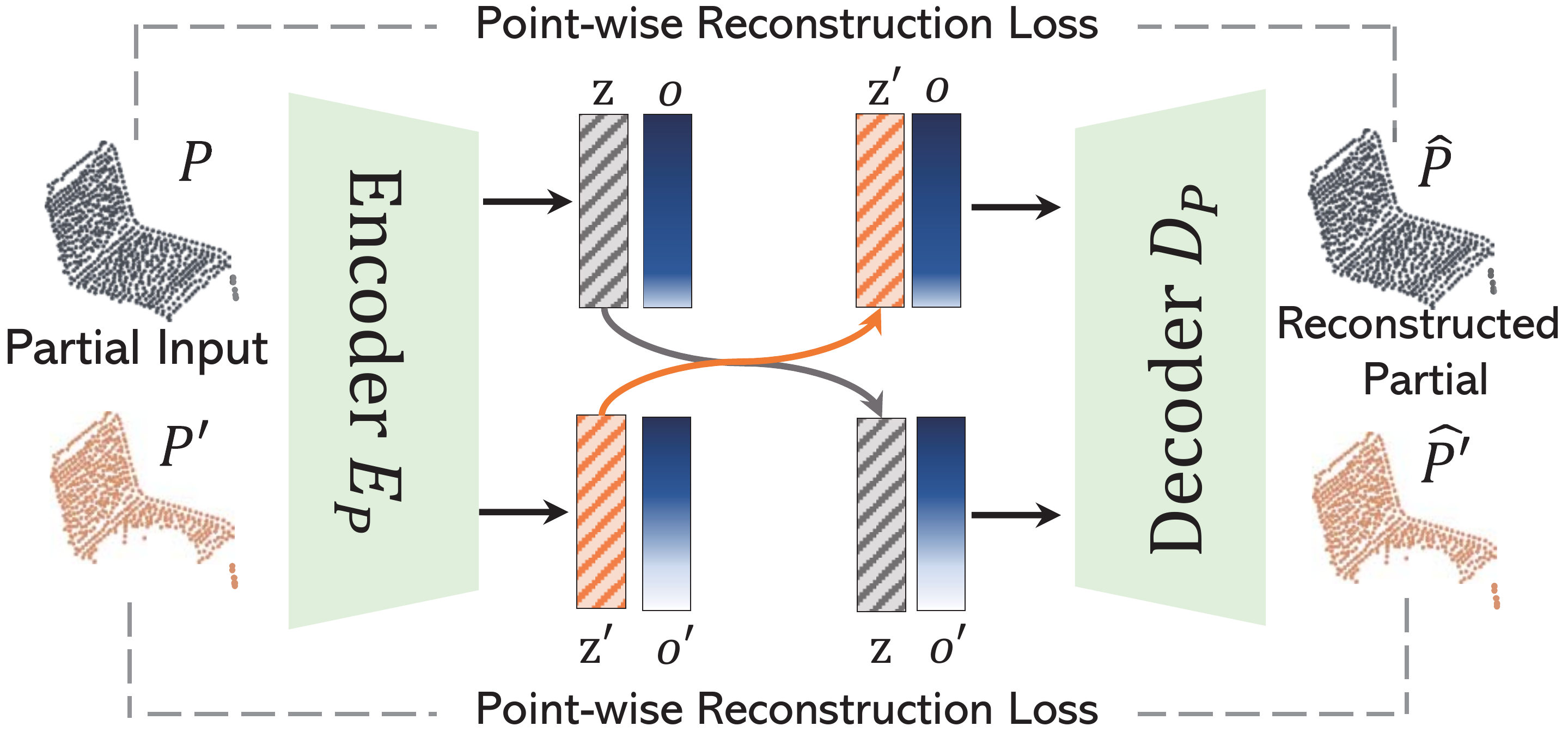}
\caption{\textbf{Illustration of the latent code swapping.~}To better decouple the complete shape and occlusion codes, we swap the complete shape codes between related partial point clouds and apply point-wise reconstruction loss to supervise the reconstructed partial point clouds. Best viewed in color.}
\label{fig:sec32_figure}
\vspace{-3mm}
\end{figure}

\noindent \textbf{Latent Code Swapping.~}
To further regularize the complete shape codes and occlusion codes, we employ a latent code swapping constraint. Specifically, as illustrated in Figure~\ref{fig:sec32_figure}, we swap the complete and occlusion codes extracted from a partial point cloud $P$ and a more occluded version of $P^{\prime}$ to reconstruct the corresponding partial point clouds. Based on our assumption on the unified space, $\mathbf{z}$ and $\mathbf{z^{\prime}}$ represent the same complete object and the partial degrees are decided by their occlusion codes. Therefore, no matter $\mathbf{o^{\prime}}$ is combined with $\mathbf{z}$ or $\mathbf{z^{\prime}}$, the same partial point cloud should be reconstructed. Therefore, we also feed the fused code from $\mathbf{z}$ and $\mathbf{o'}$ to the decoder $D_p$ and apply point-wise reconstruction loss $\mathcal{L}_{rec}$ to constrain. And the fused shape code from $\mathbf{z^{\prime}}$ and $\mathbf{o}$ is similarly processed.
Through such a latent code swapping constraint, the disentanglement of the complete shape codes and the occlusion codes are greatly improved, which leads to better shape completion.

\noindent \textbf{Latent Code Distribution.~}
In order to further constrain the reality of the complete shape codes, a shape latent code discriminator is applied to directly supervise whether the complete shape codes learned from partial point clouds match well with the real complete shape codes extracted from a complete shape auto-encoder. As illustrated in Figure~\ref{fig:framework} (c), the real shape latent code $z_c \in \mathbb{R}^d$ can be obtained from a complete point cloud auto-encoder following~\cite{wen2021cycle4completion}. The input of the auto-encoder $Y$ is a shape randomly sampled from a complete point clouds set which is not paired with $P$. The objective functions for updating latent code discriminator $\mathcal{L}_{D}^{c}$ and latent code generator $\mathcal{L}_{G}^{c}$ are similar as Eqs.~(\ref{eq:D_code}) and~(\ref{eq:G_code}) respectively.

In summary, through the constraints on the unified latent space, the complete shape and occlusion codes can be well learned to enhance the relation between the complete and partial point clouds. 

\noindent \textbf{Overall Loss.~}
The overall training objective for the two discriminators is
\begin{equation}
\mathcal{L}_{D} = \mathcal{L}_{D}^{p} + \mathcal{L}_{D}^{c}.
\end{equation}
And the overall training loss for the encoders and decoders including $E_p$, $D_p$, $E_c$ and $D_c$ is the weighted sum of point-wise reconstruction loss, structured latent supervision and adversarial losses:
\begin{equation}
\mathcal{L} = \gamma\mathcal{L}_{rec} + \beta \mathcal{L}_{z} + \mathcal{L}_{npair} + \mathcal{L}_{G}^{p} + \mathcal{L}_{G}^{c},
\end{equation}
where $\gamma$ and $\beta$ are pre-defined weight factors. 
\vspace{-0.2cm}
\section{Experiments}
\vspace{-0.1cm}
\input{tables/crn_dataset}
\input{tables/3depn_dataset}
\input{tables/partnet_dataset}
We evaluate the proposed method through extensive experiments. Besides shape completion on the virtual scan benchmarks, we also demonstrate its effectiveness compared with other methods on the widely used real-world partial scans. 

\noindent \textbf{Datasets.~}For a comprehensive comparison, we conduct experiments on both synthetic and real-world partial shapes following state-of-the-art unsupervised point cloud completion methods~\cite{chen2019unpaired,zhangunsupervised,wen2021cycle4completion,wu2020multimodal}. 
We evaluated our method on three synthetic datasets CRN~\cite{wang2020cascaded}, 3D-EPN~\cite{dai2017shape} and PartNet~\cite{mo2019partnet}, which are all derived from ShapeNet~\cite{chang2015shapenet}. For real-world scans, we evaluate on objects extracted from three datasets covering indoor and outdoor scenes, KITTI (cars)~\cite{geiger2012we}, ScanNet (chairs and tables)~\cite{dai2017scannet}, and MatterPort3D (chairs and tables)~\cite{chang2017matterport3d}.

\noindent \textbf{Evaluation Metrics.~} For datasets equipped with ground truth, we evaluate the shape completion performance using CD and F1-score following previous unsupervised point cloud completion methods~\cite{zhangunsupervised, wen2021cycle4completion,chen2019unpaired}, where F1-score is the harmonic average of the accuracy and the completeness. The Chamfer Distance is defined as:
\begin{equation}
\label{eq:cd-loss}
\begin{aligned}
\mathcal{L}_{C D}\left(\mathbf{x}_{out}, \mathbf{x}_{i n}\right) &=\frac{1}{\left|\mathbf{x}_{out}\right|} \sum_{p \in \mathbf{x}_{out}} \min _{q \in \mathbf{x}_{i n}}\|p-q\|_{2}^{2} \\
&+\frac{1}{\left|\mathbf{x}_{i n}\right|} \sum_{q \in \mathbf{x}_{i n}} \min _{p \in \mathbf{x}_{out}}\|p-q\|_{2}^{2},
\end{aligned}
\end{equation}
where $\mathbf{x}_{out}$ and $\mathbf{x}_{in}$ are two point clouds. The smaller the distance value is, the more accurate the reconstructed point cloud is.
For synthetic dataset PartNet utilized by~\cite{wu2020multimodal}, we follow the method and also use Minimum Matching Distance (MMD) metric to evaluate the accuracy of the completion. The MMD measures the quality of the completed shape and we calculate the MMD between the set of completion shapes and the set of test shapes.
For real-world scans where no ground truth is provided, we follow~\cite{xie2021style, wu2020multimodal} to evaluate the generated shapes in terms of UCD and MMD, respectively. The UCD evaluates the consistency and computes the Chamfer distance from the partial input to the predicted complete point cloud. 

\noindent \textbf{Implementation Details.~}
The proposed method follows previous unsupervised point cloud methods~\cite{chen2019unpaired,zhangunsupervised,wen2021cycle4completion,wu2020multimodal} to train single-class models separately for better fidelity. The number of points of the predicted complete shapes is 2048 for all datasets. We use 8 TITAN GPUs to implement our experiments. Specifically, we adopt an Adam optimizer with a learning rate $10^{-4}$ and a batch size of 16 per GPU to train the framework for 500 epochs. The top 5 points are kept for complete point cloud degradation. The dimension $d$ of complete shape code and occlusion code are both 96 and a number of 500 points (i.e., $K$=500) are gradually removed to generate more partial point clouds. The $\gamma$=100, $\beta$=10 and $\lambda_{gp}$=1 are set for the combination of losses.

\subsection{Completion Results on ShapeNet Benchmark}
We conduct experiments on CRN, 3D-EPN, and PartNet synthetic datasets generated from ShapeNet to demonstrate the superiority of our method over state-of-the-art unsupervised methods.

\noindent \textbf{Comparison on CRN and 3D-EPN datasets.~}
For synthetic datasets CRN and 3D-EPN equipped with ground truth, we evaluate the shape completion performance using CD and F1-score following~\cite{chen2019unpaired,zhangunsupervised,wen2021cycle4completion}.
Tables~\ref{tab:crn_dataset} and~\ref{tab:3depn_dataset} show the experiment results on the two datasets across eight categories where ``Cycle.'' and ``Inversion.'' represent Cycle4Completion and ShapeInversion respectively. As illustrated in Table~\ref{tab:crn_dataset}, the proposed method outperforms state-of-the-art unsupervised method ShapeInversion~\cite{zhangunsupervised} by large margins across most categories and achieves 12.2 CD and 85.6 F1-score surpassing~\cite{zhangunsupervised} by 2.7 and 1.7 for average CD and F1-Score metrics, respectively. For ShapeNet 3D-EPN dataset, as shown in Table~\ref{tab:3depn_dataset}, our method consistently achieves the best completion performance on most categories, especially for categories like ``chair'' and ``table'', whose shape diversity and number of training samples are relatively rich compared with other categories.
For chair and table, the CD and F1-score metrics show significant improvements (from 14.6/84.2 to 12.1/86.4 for chair and from 22.5/82.7 to 19.8/85.5 for table). There is a 0.9 gap between our method and~\cite{wen2021cycle4completion} on the car category.
Through evaluation on the two popular synthetic datasets across eight categories, our method outperforms existing methods consistently, which proves the superiority of the proposed framework that learns a unified latent space with effective and efficient structured regularization.

\noindent \textbf{Comparison on PartNet dataset.~}
We also conduct experiments on PartNet dataset utilized by MPC~\cite{wu2020multimodal}. The PartNet benchmark is generated by removing semantic parts on ShapeNet dataset. We follow~\cite{wu2020multimodal} to adopt Minimum Matching Distance on three categories to evaluate the quality of the completed shapes. As shown in Table~\ref{tab:partnet_dataset}, our method outperforms existing state-of-the-art unsupervised methods on the three categories consistently.
\begin{figure*}[t!]
\centering
\includegraphics[width=0.85\textwidth]{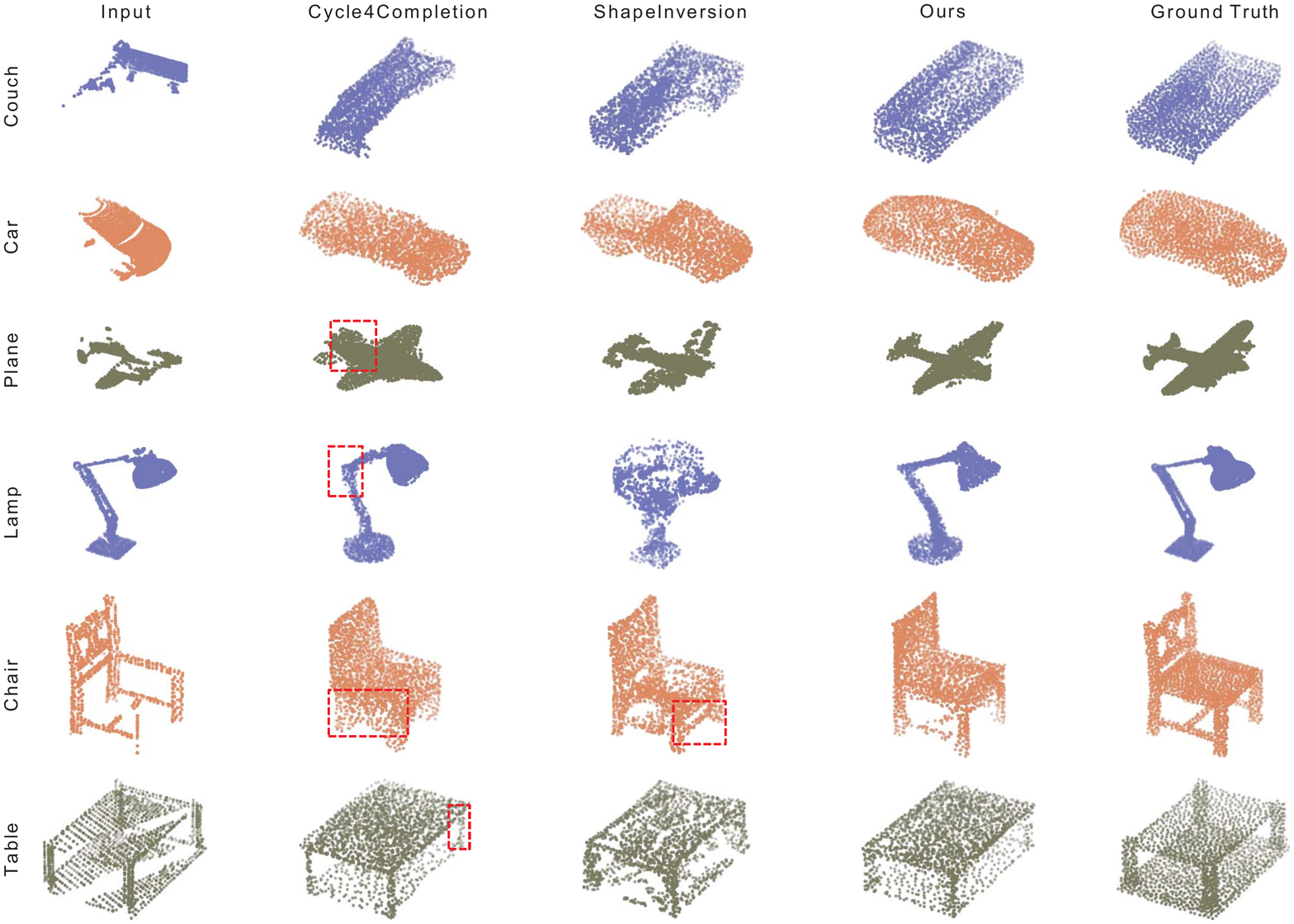}
\caption{\textbf{Point cloud completion results on ShapeNet dataset.} From left to right: partial input, results of Cycle4completion~\cite{wen2021cycle4completion}, ShapeInversion~\cite{zhangunsupervised} ours and ground truth. Our results achieve more accurate completion under severe occlusion and recover better fine-grained shape details compared with state-of-the-art methods. Better viewed in color and zoom in.}
\label{fig:shapenet_vis}
\vspace{-6mm}
\end{figure*}
 
\noindent \textbf{Qualitative Results.~}Figure~\ref{fig:shapenet_vis} illustrates the qualitative results of the same samples from the ShapeNet dataset.
Despite the efforts of previous approaches, they usually fail to deal with severe occlusion cases and can not maintain shape details.
As shown by the couch and car in Figure~\ref{fig:shapenet_vis} first two rows, in the case of severe occlusion, the completed point clouds of previous methods do not represent the target objects (see the large missing regions which are not recovered correctly). However, our method can accurately recover the complete point cloud of the target object, even when only fairly limited information is available under severe occlusion.
What's more, our method reconstructs more accurate complete point clouds equipped with better fine-grained shape details. As shown by the red dotted boxes, our method can generate more accurate complete shapes at the corner of the lamps, the tail of the planes, and the legs of chairs and tables. We attribute the great results to the learned unified latent space and properly applied structured latent supervisions, which leads to more reasonable predicted complete point clouds and better consistency between partial and complete point clouds.
\subsection{Completion Results on Real-World Scans}
\vspace{-0.1cm}
We investigate the generalization of the proposed method on various real-world datasets including both outdoor and indoor scenes, where the objects tend to be more incomplete and noisier. The trained car, chair and table models on CRN dataset are directly utilized to predict complete point clouds on KITTI, ScanNet and MatterPort3D datasets without any further fine-tuning process. As shown in Table~\ref{tab:real-ucd} (Cycle.~\cite{wen2021cycle4completion} \textit{vs.} Ours), our method significantly outperforms Cycle4Completion~\cite{wen2021cycle4completion} across multiple categories on all the three real-scan datasets.
For the comparison with ShapeInversion~\cite{zhangunsupervised}, as the inversion process is to minimize the UCD loss directly between partial-complete pairs, it is unfair to compare our method that does not involve GAN inversion with ShapeInversion~\cite{zhangunsupervised}. However, our method is also compatible with ShapeInversion~\cite{zhangunsupervised}. When integrating GAN inversion on the top of our method, it can surpass ShapeInversion on all various real-world scans, as shown in Table~\ref{tab:real-ucd} (Inversion.~\cite{zhangunsupervised} \textit{vs.} Ours + \textit{Inversion}), which demonstrates that our method can enhance consistency between the predicted complete point cloud and the partial input. 
On the other hand, the results of our method in Table~\ref{tab:real-mmd} also surpass other unsupervised methods consistently. 

\begin{figure}[t!]
\centering
\includegraphics[width=\linewidth]{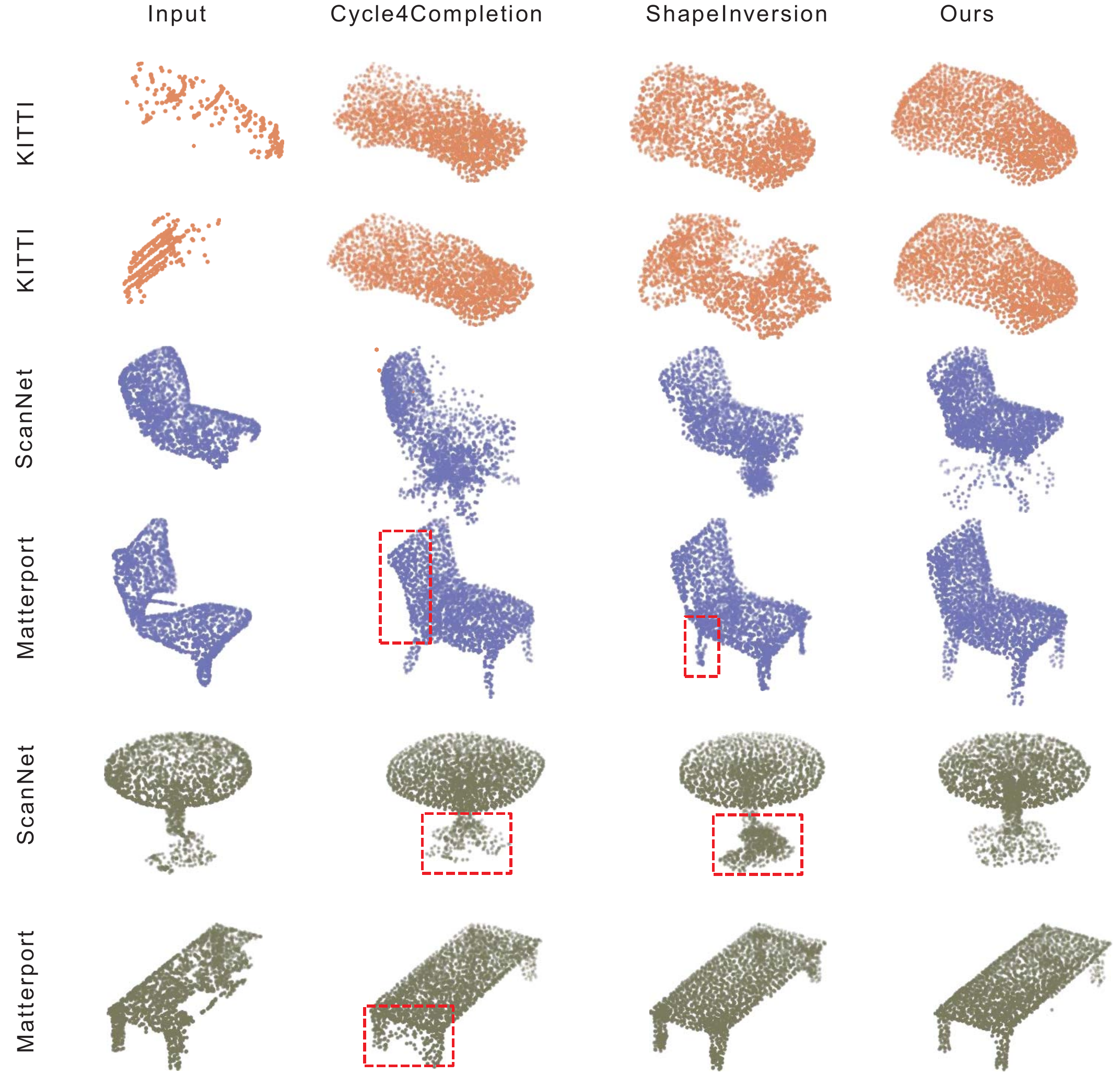}
\vspace{-3mm}
\caption{\textbf{Point cloud completion results on real-world scans.} From left to right: partial input, results of Cycle4completion~\cite{wen2021cycle4completion}, ShapeInversion~\cite{zhangunsupervised} and ours, respectively. Our method predicts more reasonable complete shape results compared with other state-of-the-art methods. Better viewed in color and zoom in.}
\label{fig:real_vis}
\vspace{-5mm}
\end{figure}

In addition, as shown in Tables~\ref{tab:real-ucd} and~\ref{tab:real-mmd}, we also compare the generalization of our models with those of state-of-the-art fully-supervised methods~\cite{xie2020grnet,yu2021pointr} on the real scans. Their authors' official released models are used here.
Our unsupervised model can outperform them on multiple categories of real scans, which demonstrates that the proposed unsupervised method has better generalization ability on real-world scans than the supervised methods, which are specifically trained to only fit their original synthetic data.
Figure~\ref{fig:real_vis} shows the completion results of our method on real data, which indicates that even under severe occlusions (such as KITTI's car), our method can still generate reasonable complete shapes. 
\input{tables/real_dataset_ucd}
\input{tables/real_dataset_mmd}
\subsection{Ablation Study}
\vspace{-0.1cm}
To verify the effectiveness of each component of the proposed method, we conduct a series of experiments on four representative categories on ShapeNet CRN dataset.

\noindent \textbf{Effect of Unified Latent Space for Point Clouds Encoding.~}To evaluate the benefits of introducing the unified latent code space for unsupervised point cloud completion, we compare two alternative strategies that do not encode occlusion codes as soft weighting vectors. Instead of fusing the codes via element-wise multiplication, we test fusing the shape and occlusion codes via concatenation or element-wise addition. Table~\ref{tab:ab-fusion} shows the quantitative results of the compared schemes. 
We employ our simplified model, which multiples the shape and occlusion codes, and has point and code discriminators and code swapping constraints under our unified space design but without using the ranking constraints as the baseline (denoted as ``Uni. Space'') in Table~\ref{tab:ab-fusion}.
The average CD drops from 18.3 to 19.1 and 18.6 when fusing the two codes via concatenation or addition, which proves that the learned unified latent space is conducive to unsupervised point cloud completion, and also creates a foundation for integrating our stronger ranking supervision.
\input{tables/ab_fusion}

\noindent \textbf{Effect of Structured Ranking Supervision.~}To evaluate the effect of our proposed structured ranking supervision, we adopt different ranking supervisions, N-pair loss and triplet loss with varying hyper-parameters.
As shown in Table~\ref{tab:ab-ranking_loss}, our model with the triplet loss with different hyper-parameters (denoted as ``+triplet'') shows better performance compared with not using the ranking supervisions. Specifically the average Chamfer Distance improves from 18.3 to 17.2. 
Furthermore, when equipped with N-pair loss, the results are further improved and the average CD has an obvious improvement from 18.3 to 15.4. It demonstrates that the proposed structured ranking supervisions assist the learning of the unified latent space. Therefore, for our full model, we use N-pair loss without any hyper-parameter as our final structured ranking supervision.
\input{tables/ab_structured_loss}

\noindent \textbf{Effect of Discriminators.~}We also test the effects of the two discriminators used in our system via deducting the code discriminator (denoted as w/o codeD) or point cloud discriminator (denoted as w/o pointD) from our final model to compare with our full solution (denoted as Full Model). As shown in Table~\ref{tab:ab-discriminator}, when removing the point cloud discriminator or latent code discriminator, the average Chamfer Distances drops 4 and 5.4 points, respectively. It proves that the integration of the discriminators can assist the complete model to generate better predictions via providing direct guidance about the distributions of complete shape latent codes and point clouds.
\input{tables/ab_A_woDiscriminator}

\noindent \textbf{Effect of Latent Code Swapping.~}The occlusion codes swapping between a series of related partial point clouds is designed to improve the disentanglement of complete shape and occlusion codes. Here we test removing the design of swapping occlusion code for reconstructing pairs of partial point clouds (denoted as w/o code swap) to compare with our solution. As shown in the last row of Table~\ref{tab:ab-discriminator}, without latent codes swapping, the system shows much worse performance on all categories, decreasing for 2.3 CD from 15.4 to 17.7. It demonstrates that generating related partial point clouds and swapping their latent codes provide effective supervisions on decoupling the complete shape and occlusion codes more thoroughly.
\vspace{-0.2cm}
\section{Limitations and Conclusion}
\vspace{-0.2cm}
In this paper, we propose to learn a unified and structured latent space, which encodes both partial and complete point clouds to improve partial-complete geometry consistency under an unsupervised manner. Furthermore, we apply tailored structured latent supervisions between a series of related partial point clouds to enhance the learning of the structured latent space. Extensive experiments prove that the proposed method consistently achieves state-of-the-art performance on both synthetic and real-world benchmarks. 

Although our method has made great progress in accuracy, there are still limitations for some fine-grained structure reconstruction of objects, such as the complex texture structure of chairs. These limitations may be solved by designing better decoders or introducing implicit functions.

\noindent \textbf{Acknowledgments:~}This work is supported in part by Centre for Perceptual and Interactive Intelligence Limited, in part by the General Research Fund through the Research Grants Council of Hong Kong under Grants (Nos. 14204021, 14207319, 14203118, 14208619), in part by Research Impact Fund Grant No. R5001-18, in part by CUHK Strategic Fund.

\clearpage

{\small
\bibliographystyle{ieee_fullname}
\bibliography{egbib}
}
\end{document}

%% file: tables/crn_dataset.tex
\begin{table*}[t!]
\caption{Shape completion performance on CRN benchmark. The numbers shown are [CD$\downarrow$ /F1$\uparrow$], where CD is scaled by $10^4$.}
\vspace{-3mm}
 \scalebox{0.97}{
\label{tab:crn_dataset}
\begin{tabular}{l|c|c|c|c|c|c|c|c|c}
\hline 
Methods & Plane    & Cabinet   & Car       & Chair     & Lamp      & Sofa      & Table     & Boat      & Average   \\ \hline
Pcl2pcl~\cite{chen2019unpaired} & 9.7/89.1 & 27.1/68.4 & 15.8/80.8 & 26.9/70.4 & 25.7/70.4 & 34.1/58.4 & 23.6/79.0 & 15.7/77.8 & 22.4/74.2 \\ 
Cycle.~\cite{wen2021cycle4completion} &  5.2/94.0 & 14.7/82.1 & 12.4/82.1& 18.0/77.5 & 17.3/77.4 & 21.0/75.2 & 18.9/81.2 & 11.5/84.8 & 14.9/81.8 \\ 
Inversion.~\cite{zhangunsupervised}   &5.6/94.3 & 16.1/77.2 & 13.0/85.8 & 15.4/81.2 & 18.0/\textbf{81.7} & 24.6/78.4 & \textbf{16.2/85.5} & 10.1/87.0 & 14.9/83.9 \\ 
Ours & \textbf{3.9/95.9} & \textbf{13.5/83.3} & \textbf{8.7/90.4} & \textbf{13.9/82.3}  & \textbf{15.8}/81.0 & \textbf{14.8/81.6} & 17.1/82.6 & \textbf{10.0/87.6} & \textbf{12.2/85.6} \\ \hline
\end{tabular}}
\end{table*}

%% file: tables/3depn_dataset.tex
\begin{table*}[t!]
\caption{Shape completion performance on 3D-EPN benchmark. The numbers shown are [CD$\downarrow$ /F1$\uparrow$], where CD is scaled by $10^4$.}
\vspace{-3mm}
\scalebox{0.97}{
\label{tab:3depn_dataset}
\begin{tabular}{l|c|c|c|c|c|c|c|c|c}
\hline Methods & Plane    & Cabinet   & Car       & Chair     & Lamp      & Sofa      & Table     & Boat      & Average   \\ \hline
Pcl2pcl~\cite{chen2019unpaired} & 4.0/-- & 19.0/-- & 10.0/-- & 20.0/-- & 23.0/-- &  26.0/-- & 26.0/-- & 11.0/-- & 17.4/-- \\ 
Cycle.~\cite{wen2021cycle4completion} & 3.7/96.4 & 12.6/\textbf{87.1} &    \textbf{8.1/91.8} &   14.6/84.2 &  18.2/80.6 &  26.2/71.7 & 22.5/82.7 &  8.7/89.8 & 14.3/85.5 \\ 
Inversion.~\cite{zhangunsupervised} & 4.3/96.2  & 20.7/79.4 &11.9/86.0& 20.6/81.1&  25.9/78.4  & 54.8/74.7 & 38.0/80.2 & 12.8/85.2 &  23.6/82.7    \\ 

Ours  & \textbf{3.5/96.8}   & \textbf{12.2}/86.4  & 9.0/88.4 & \textbf{12.1/86.4} & \textbf{17.6/81.6}  & \textbf{26.0/75.5}  & \textbf{19.8/85.5} &  \textbf{8.6/89.8}   &  \textbf{13.6/86.3}    \\ \hline
\end{tabular}}
\vspace{-3mm}
\end{table*}

%% file: tables/partnet_dataset.tex
\begin{table}[t!]
\caption{Shape completion performance on PartNet benchmark. We evaluate the results with MMD$\downarrow$, which is scaled by $10^2$.}
\label{tab:partnet_dataset}
\begin{tabular}{l|c|c|c|c}
\hline Methods & Chair & Lamp & Table & Average  \\ \hline
Pcl2pcl~\cite{chen2019unpaired}& 1.90 & 2.50 & 1.90  &  2.10  \\ 
MPC~\cite{wu2020multimodal}& 1.52 & 1.97 & 1.46  &  1.65  \\ 
Cycle.~\cite{wen2021cycle4completion}  & 1.71  & 3.46 & 1.56 & 2.24\\ 
Inversion.~\cite{zhangunsupervised} & 1.68 & 2.54 & 1.74 & 1.98 \\ 
Ours & \textbf{1.43}  & \textbf{1.95} & \textbf{1.37} & \textbf{1.58}      \\ \hline
\end{tabular}
\vspace{-5mm}
\end{table}

%% file: tables/real_dataset_ucd.tex
\begin{table}[t!]
\caption{Shape completion performance on the real scans. The results are evaluated by UCD$\downarrow$, where UCD is scaled by $10^4$. \textit{sup.}: supervised methods.}
\vspace{-2mm}
\label{tab:real-ucd}
\scalebox{0.85}{
\begin{tabular}{l|c|c|c|c|c|c}
\hline
\multirow{2}{*}{Methods} & \multirow{2}{*}{\textit{sup.}} & \multicolumn{2}{c|}{ScanNet} & \multicolumn{2}{c|}{MatterPort3D} & KITTI \\ \cline{3-7}
& & Chair & Table & Chair & Table & Car \\ \hline
GRNet~\cite{xie2020grnet} & yes & 1.6 & 1.6 & 1.6 & 1.5 & 2.2 \\
PoinTr~\cite{yu2021pointr} & yes & 1.7 & 1.5 & 1.8 & 1.3 & 1.9 \\ \hline
Pcl2pcl~\cite{chen2019unpaired} & no & 17.3 & 9.1 & 15.9 & 6.0 & 9.2 \\ 
Cycle.~\cite{wen2021cycle4completion} & no & 9.4 & 4.3 & 4.9 & 4.9 & 9.4  \\ 
Inversion.~\cite{zhangunsupervised} & no & 3.2 & 3.3 & 3.6 & 3.1 & 2.9   \\ 
Ours & no & 3.2 & 2.7 & 3.3 & 2.7 & 4.2  \\ 
Ours + \textit{Inversion} & no & \textbf{1.1} & \textbf{0.87} & \textbf{1.1} & \textbf{0.87} & \textbf{0.76} \\ \hline
\end{tabular}}
\vspace{-6mm}
\end{table}

%% file: tables/real_dataset_mmd.tex
\begin{table}[t!]
\caption{Shape completion performance on the real scans. We evaluate the results with MMD$\downarrow$, where MMD is scaled by $10^2$. \textit{sup.}: supervised methods.}
\label{tab:real-mmd}
\vspace{-2mm}
\scalebox{0.85}{
\begin{tabular}{l|c|c|c|c|c|c}
\hline
\multirow{2}{*}{Methods} & \multirow{2}{*}{\textit{sup.}} & \multicolumn{2}{c|}{ScanNet}    & \multicolumn{2}{c|}{MatterPort3D} & KITTI  \\ \cline{3-7} 
& & Chair          & Table          & Chair           & Table           & Car             \\ \hline
GRNet~\cite{xie2020grnet} & yes & 6.070 & 6.302 & 6.147 & 6.911 & 2.845 \\
PoinTr~\cite{yu2021pointr} & yes & 6.001 & 6.089 &6.248 & 6.648 & 2.790 \\ \hline
Cycle.~\cite{wen2021cycle4completion} & no & 6.278 & 5.727 & 6.022 & 6.535 & 3.033 \\
Inversion.~\cite{zhangunsupervised} & no & 6.370  & 6.222& 6.360 & 7.110 & 2.850  \\       
Ours & no  & \textbf{5.893} & \textbf{5.541} & \textbf{5.770}  & \textbf{6.076}  & \textbf{2.742} \\ \hline
\end{tabular}}
\vspace{-5mm}
\end{table}

%% file: tables/ab_fusion.tex
\begin{table}[t!]
\caption{Comparison of different schemes for fusing complete shape and occlusion codes. CD$\downarrow$ scaled by $10^4$ are reported here.}
\vspace{-2mm}
\label{tab:ab-fusion}
\scalebox{0.90}{
\begin{tabular}{l|c|c|c|c|c}
\hline 
Method & Chair & Lamp & Sofa & Table & Avg.  \\ \hline
Uni. Space &\textbf{16.2} & 20.3 & \textbf{16.1} & \textbf{20.5}  & \textbf{18.3} \\ \hline
Concatenation & 17.2     &   20.7   &  17.5    &  20.8  & 19.1 \\
Addition & 17.2   &  \textbf{20.1}   &  16.3    &   20.8    &  18.6 \\ 
\hline
\end{tabular}}
\vspace{-2mm}
\end{table}

%% file: tables/ab_structured_loss.tex
\begin{table}[t!]
\caption{Effect of different ranking supervisions. CD$\downarrow$ scaled by $10^4$ are reported here.}
\label{tab:ab-ranking_loss}
\vspace{-2mm}
\scalebox{0.90}{
\begin{tabular}{l|c|c|c|c|c}
\hline 
Method & Chair & Lamp & Sofa & Table & Avg.\\ \hline
Uni. Space & 16.2	& 20.3	& 16.1	& 20.5  & 18.3\\ \hline
+triplet& \makecell{14.7 \\ ($\delta=5$)} & \makecell{19.5 \\ ($\delta=2$)} & \makecell{15.0 \\ ($\delta=10$)} & \makecell{19.5 \\ ($\delta=5$)} &17.2 \\
\hline
+n-pair & \textbf{13.9} & \textbf{15.8} & \textbf{14.8} & \textbf{17.1}  & \textbf{15.4}\\ \hline
\end{tabular}}
\vspace{-5mm}
\end{table}

%% file: tables/ab_A_woDiscriminator.tex
\begin{table}[t!]
\caption{Effect of discriminators and latent code swapping. CD$\downarrow$ scaled by $10^4$ are reported here.}
\vspace{-2mm}
\label{tab:ab-discriminator}
\begin{tabular}{l|c|c|c|c|c}
\hline 
Method & Chair & Lamp & Sofa & Table & Avg.  \\ \hline
Full Model & \textbf{13.9} & \textbf{15.8} & \textbf{14.8} & \textbf{17.1}  & \textbf{15.4}\\ \hline
w/o pointD & 14.7 & 23.6 & 20.0 & 19.4& 19.4\\
w/o codeD & 18.5 & 21.7 & 22.0 &20.8 & 20.8\\ 
w/o code swap  & 14.7 & 20.0 & 17.0& 19.2 & 17.7\\ \hline
\end{tabular}
\vspace{-5mm}
\end{table}